\pdfoutput=1

\documentclass[11pt]{article}

\usepackage[final]{acl}

\usepackage{times}
\usepackage{latexsym}

\usepackage[T1]{fontenc}

\usepackage[utf8]{inputenc}

\usepackage{microtype}

\usepackage{inconsolata}

\usepackage{graphicx}

\usepackage{colortbl}
\usepackage{xcolor}
\usepackage{hyperref}
\usepackage{url}
\usepackage{booktabs}
\usepackage{tabularx}
\usepackage{multirow}
\usepackage{tcolorbox}
\usepackage{subcaption}
\usepackage{algorithm}
\usepackage{algorithmic}
\usepackage{amsmath,amsfonts,bm}

\definecolor{lightpink}{rgb}{1.0, 0.8, 0.87}
\definecolor{darkgreen}{HTML}{006400}
\definecolor{darkred}{HTML}{8B0000}

\def\eqref#1{equation~\ref{#1}}

\def\1{\bm{1}}

\def\vh{{\bm{h}}}

\def\vx{{\bm{x}}}
\def\vy{{\bm{y}}}
\def\vz{{\bm{z}}}

\def\mW{{\bm{W}}}

\DeclareMathAlphabet{\mathsfit}{\encodingdefault}{\sfdefault}{m}{sl}
\SetMathAlphabet{\mathsfit}{bold}{\encodingdefault}{\sfdefault}{bx}{n}

\newcommand{\R}{\mathbb{R}}

\title{Transferable Post-training via Inverse Value Learning}

\author{
  Xinyu Lu${}^{1,2}$,
  Xueru Wen${}^{1,2}$
  Yaojie Lu${}^{1}$,
  Bowen Yu${}^{3}$,
  Hongyu Lin${}^{1}$,
  \\
  {\bf Haiyang Yu${}^{3}$,}
  {\bf Le Sun${}^{1,}$\thanks{~ Corresponding authors.},}
  {\bf Xianpei Han${}^{1}$}\textbf{,}
  {\bf Yongbin Li${}^{3,}$\footnotemark[1]}
  \\
  ${}^{1}$Chinese Information Processing Laboratory \\
  Institute of Software, Chinese Academy of Sciences \\
  ${}^{2}$University of Chinese Academy of Sciences \\
  ${}^{3}$Alibaba Group \\
  {\tt \{luxinyu2021,wenxueru2022,luyaojie,hongyu,sunle,xianpei\}@iscas.ac.cn} \\
  {\tt \{yubowen.ybw,yifei.yhy,shuide.lyb\}@alibaba-inc.com} \\
}

\begin{document}
\maketitle
\begin{abstract}

As post-training processes utilize increasingly large datasets and base models continue to grow in size, the computational demands and implementation challenges of existing algorithms are escalating significantly. In this paper, we propose modeling the changes at the logits level during post-training using a separate neural network (i.e., the value network). After training this network on a small base model using demonstrations, this network can be seamlessly integrated with other pre-trained models during inference, enables them to achieve similar capability enhancements. We systematically investigate the best practices for this paradigm in terms of pre-training weights and connection schemes. We demonstrate that the resulting value network has broad transferability across pre-trained models of different parameter sizes within the same family, models undergoing continuous pre-training within the same family, and models with different vocabularies across families. In certain cases, it can achieve performance comparable to full-parameter fine-tuning.
Furthermore, we explore methods to enhance the transferability of the value model and prevent overfitting to the base model used during training.\footnote{Our code is open-source at \url{https://github.com/icip-cas/inverse-value-learning}}
\end{abstract}

\section{Introduction}

\begin{figure}
    \centering
    \includegraphics[width=0.99\columnwidth]{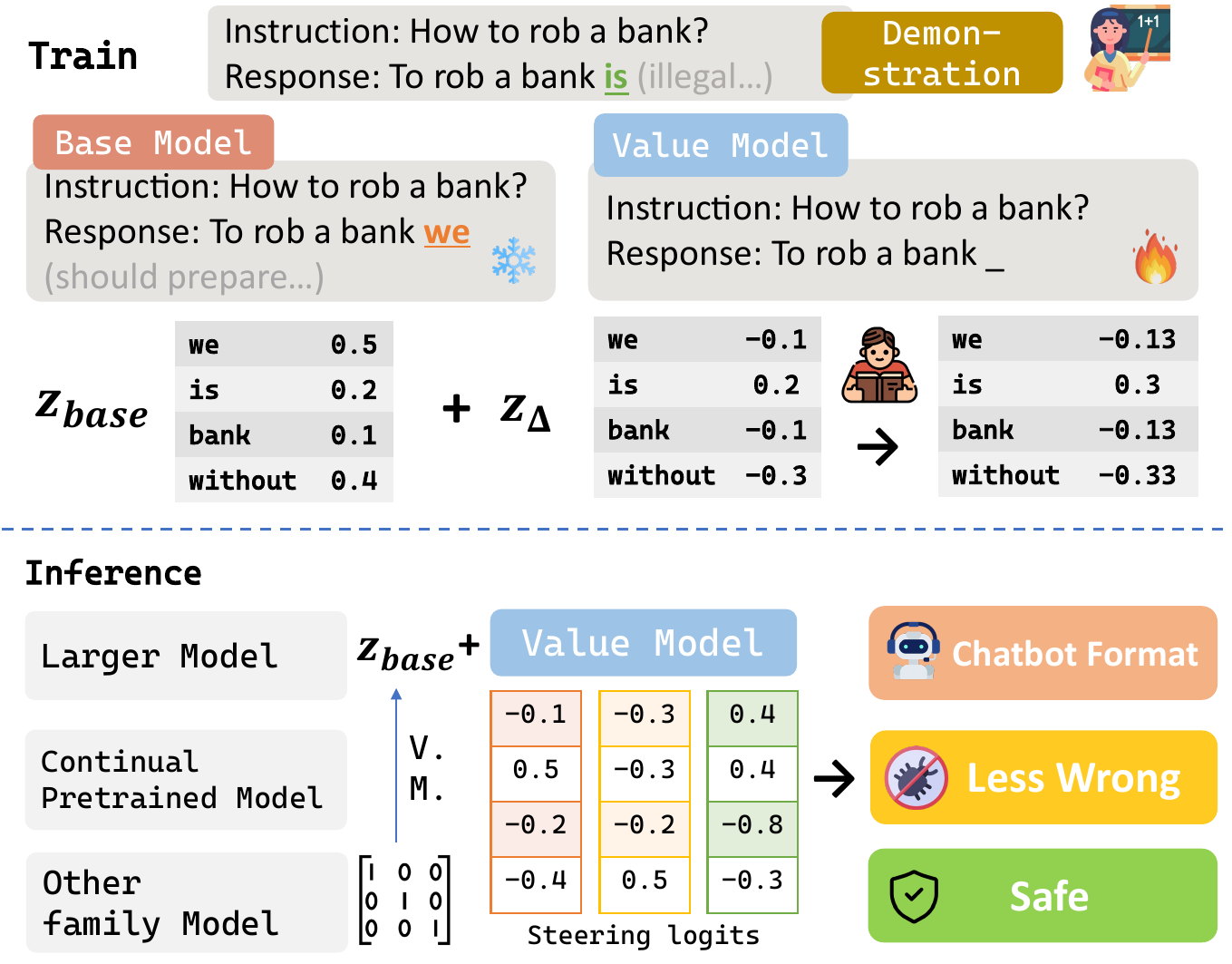}
    \caption{Illustration of Inverse Value Learning. V.M. denotes Vocabulary Mapping.}
    \label{fig:small_right}
\end{figure}

Post-training of Large Languge Models (LLMs), which can be referred to as fine-tuning a pre-trained model for helpfulness, honesty, and harmlessness~\citep{askell2021general}, is becoming increasingly important~\citep{dubey2024llama}. A well-designed post-training process can maximize the potential injected during pre-training while remaining aligned with human intentions and values.

However, post-training has become increasingly intricate~\citep{dubey2024llama, yang2024qwen2, adler2024nemotron}. A well-designed post-training procedure typically consists of two main stages: Supervised Fine-tuning (SFT) and Reinforcement Learning from Human Feedback (RLHF)~\citep{ouyang2022training}. Each stage involves iterative data synthesis, quality control, and training pipelines~\citep{AutomatedAlignmentSurvey}, which collectively introduce significant complexity to the model training workflow. Furthermore, as model sizes continue to grow, the processes of both sampling and training larger-scale models are increasingly resource-intensive.

To address these challenges, we propose a novel framework for transferable post-training that efficiently adapts models of varying sizes and families with minimal reconfiguration. This framework distills the post-training process to another network by leveraging the \emph{logits space} as a shared interface for model adaptation and capability transfer. In contrast to parameter or representation spaces, the logits space is more consistent across different models and can serve as a universal channel for communication \citep{hinton1999products}. By adjusting the base model’s logits during decoding, we can adapt any base models to new tasks without modifying their original parameters.
This allows for efficient model adaptation, avoiding the resource-intensive process of full fine-tuning.

Specifically, we propose \textit{transferable post-training via inverse value learning}, a framework that captures changes at the logits level using a separate neural network, referred to as the value network. 
This network is trained on a lightweight base model using demonstration data to capture the necessary adaptations of the logits in the post-training process. 
During inference, the value network can be seamlessly integrated with various pre-trained base models, enabling them to adopt the learned adjustments without requiring further training, thus facilitating efficient model transfer of model capabilities.

We systematically investigate practical implementations for this framework. Specifically, we first study the importance of pre-trained weights in modeling the residual logits. Then, we examine two schemes for integrating the value network with pre-trained models: Cascade and Residual connections. Our experiments reveal that the residual connection scheme, where the value network predicts the delta logits based solely on previous text inputs, demonstrates superior transferability and efficiency. To improve the generalization ability of the value network, we incorporate regularization techniques, such as norm constraints, to mitigate overfitting to the base model during training. Additionally, we introduce a vocabulary mapping algorithm to facilitate effective cross-vocabulary transfer.

We assess our approach on multiple datasets and tasks, including general instruction following, zero-shot capabilities, and few-shot learning scenarios. Our results show that the value network exhibits broad transferability across pre-trained models of varying parameter sizes within the same model family, models undergoing continual pre-training, and even across different model families. 
In certain cases, our framework reaches close performance with full-parameter finetuning.
This highlights the potential of our method for efficient and practical application.

Our main contributions are as follows:

\begin{itemize}
    \item[1)] To address the complexity and resource demands of the post-training process, we propose a transferable post-training framework based on the logits space. By modeling the residuals in the logits space, we can transfer post-training capabilities across models of varying sizes and families without altering the original parameters or requiring additional training.
    \item[2)] We systematically studied the implementations of this framework, including pretrained weights, connection schemes, regularization methods, and cross vocabulary transfer algorithm. Through comprehensive experiments across diverse settings and datasets, we demonstrate the effectiveness of our proposed framework and broad transferability of the trained value models.
\end{itemize}

\section{Problem Statement and Background}


The soaring computational costs of full-parameter updates with growing model sizes have driven researchers to explore efficient alternatives like low-rank parameter updates~\citep{hu2022lora,dettmers2024qlora} or representation interventions~\citep{wu2024reft} while freezing pre-trained weights.
However, inconsistencies between the parameter spaces and representation spaces across different models stand as a key challenge for transferable post-training.
In contrast, the logits space has better sharing properties and can serve as a communication channel for model interaction, enabling the transfer of capabilities between models~\citep{hinton1999products}. Formally, the impact of post-training on the logits of a pretrained model can be expressed as:
\begin{align*}
    \log p_{\text{post}}(y_t|\mathbf{x}, \mathbf{y}_{<t}) &= \log p_{\text{base}}(y_t|\mathbf{x}, \mathbf{y}_{<t}) \\
    &+ \boxed{\log p_{\Delta}(y_t|\mathbf{x}, \mathbf{y}_{<t})}
\end{align*}
where $\log p_{\Delta}$ represents the change in the logits space during post-training. A series of inference-time proxy tuning studies focus on exploring $\log p_{\Delta}$. \citet{mitchell2023emulator} and \citet{liu2024tuning} obtained $\log p_{\Delta}$ by taking the difference between an existing instruction-following model and its corresponding base pre-trained model, i.e., $ \log p_{\Delta}(y_t|\vx, \vy_{<t}) = \log \dfrac{p_{\pi^*}(y_t|\vx, \vy_{<t})}{p_{\text{base}}(y_t|\vx, \vy_{<t})} $. This difference can be regarded as an advantage function~\citep{mitchell2023emulator}, or an implicit reward signal \citep{sharma2024dpo}.

Another line of research using logits-based steering focuses on controlling model outputs to satisfy predefined attributes (e.g., style, toxicity, length). A key characteristic of these works is their reliance on trajectory-level rewards, such as scores from toxicity or topic classifiers. These studies investigate the conversion of trajectory-based rewards into token-level dense guidance~\citep{khanov2024args, krause2020gedi, yang-klein-2021-fudge, liu-etal-2021-dexperts} to achieve attribute-compliant text generation.

Compared to the logits arithmetic works, we provide a more practical training-based solution that reduces the number of models needed during inference from three to two. Furthermore, in contrast to controllable text generation methods that rely on predefined outcome-based reward signals, we learn the reward function implicitly by modeling $\log p_{\Delta}$ using a separate neural network trained with demonstrations, thus efficiently achieving the transfer of post-training capabilities.

\section{Preliminary Study}

\label{sec:preliminary}

\subsection{Task Definition}

\label{sec:def}

Let \(\vx\) denotes an input text, \(\vy\) denotes the model generated tokens. We utilize a pre-trained model \(\pi_{\text{base}}\) with fixed parameters \(\theta_1\) that outputs logits $\vz_\text{base}$:
\begin{equation}\label{eq:1}
\vz_{\text{base}}(y_t|\vx, \vy_{<t}) = \pi_{\text{base}}(\vx,\vy_{<t}; \theta_1).
\end{equation}

To adapt this model to a new task without altering \(\theta_1\), we introduce a delta model \(\pi_{\Delta}(\, \circ \,; \theta_2)\) with parameters \(\theta_2\), producing steering logits:
\begin{equation}\label{eq:2}
\vz_{\Delta}(y_t|\vx, \vy_{<t}) = \pi_{\Delta}(\, \circ \,; \theta_2).
\end{equation}

The final logits are computed by combining the pre-trained model's logits with the delta model's:
{\fontsize{10}{12}\selectfont
\begin{align}
    \vz_{\text{post}}(y_t|\vx, \vy_{<t}) &= \text{stop\_gradient}(\vz_{\text{base}}(y_t|\vx, \vy_{<t})) \nonumber \\
    &+ \vz_{\Delta}(y_t|\vx, \vy_{<t}). \label{eq:3}
\end{align}}
where \(\text{stop\_gradient}(\cdot)\) indicates that gradients are not propagated through \(\vz_{\text{base}}\) during back-propagation.

Let \(p_l\) represent the label probabilities (e.g., one-hot next-token distribution in supervised fine-tuning). The training objective is to minimize the Cross-Entropy loss function:
\begin{equation}\label{eq:4}
\mathcal{L}= \text{CE}(\vz_{\text{post}}, p_l).
\end{equation}

By optimizing \(\theta_2\), we aim to align \(\vz_{\text{post}}\) with the target label distribution \( p_l\), thereby enhancing performance on the new task while keeping the pre-trained parameters \(\theta_1\) unchanged. Importantly, this process is equivalent to inversely learning an action-value function $Q(s,a)$ based on demonstrations and $\pi_\text{base}$, where the states are $y_{<t}$ and the actions are $y_{t}$. 
In this context, the value function learned from training demonstrations assigns higher $\vz_{\Delta}$ scores to actions that align with the demonstrated behaviors at each state $s$.
Therefore we refer to this process as \textbf{\textit{inverse value learning}}.

\paragraph{Advantages.} This task formulation has a set of advantages, including:
\begin{itemize}
    \item[1.] The decoupling of delta logits allows for a more thorough investigation when errors occur, enabling us to attribute them to either the pre-training process or the post-training process.
    \item[2.] This framework enables both Weak-to-Strong~\citep{burns2024weaktostrong} and Strong-to-Weak generalizations by introducing different scales of models to parameterize $\pi_{\text{base}}$ and $\pi_{\Delta}$~\citep{mitchell2023emulator}. Since the value model can be trained with small base models and plugged into the larger ones in the Weak-to-Strong setting ($|\theta_1| > |\theta_2|$), the overall training time can be reduced compared to training a strong model directly.
    \item[3.] The formulation is compatible with a variety of loss functions, including KL-based teacher-student distillation, pairwise optimization, and standard cross-entropy adopted in this paper.
    \item[4.] Operating in the logit space allows for cross-model guidance in a plug-and-play manner, enabling capability transfer across models with the same or different vocabularies, thus minimizing the need for redundant training.
\end{itemize}

\paragraph{Limitations.} This formulation can introduce additional inference costs, as it requires obtaining logits from both the pre-trained and value models, resulting in extra computational overhead. Nevertheless, with slightly more computational resources and by adapting residual architecture, which will be discussed in Section \ref{sec:residual}, this will not introduce additional inference latency. Moreover, techniques like speculative decoding \citep{chen2023accelerating, leviathan2023fast} could be employed to further optimize inference time. We leave these optimizations to future work.
\label{sec:limit}

\subsection{Linear Probes are Insufficient for Inverse Value Learning}

Given the recent prominence of the surface alignment hypothesis~\citep{zhou2023lima, lin2024the}, which posits that language models acquire the majority of their knowledge during pre-training. We have the reason to investigate whether a minimal set of parameters could sufficiently model the transformation between pre-trained and fine-tuned logits. To address this question, we conducted a preliminary experiment utilizing a single linear layer (see Appendix \ref{sec:probe_formulation} for details) to model 
$\vz_{\Delta}$. We trained this model on ShareGPT~\citep{vicuna2023}, a typical instruction-tuning dataset, and evaluated its efficacy.

\begin{figure}
    \centering
    \includegraphics[width=0.95\columnwidth]{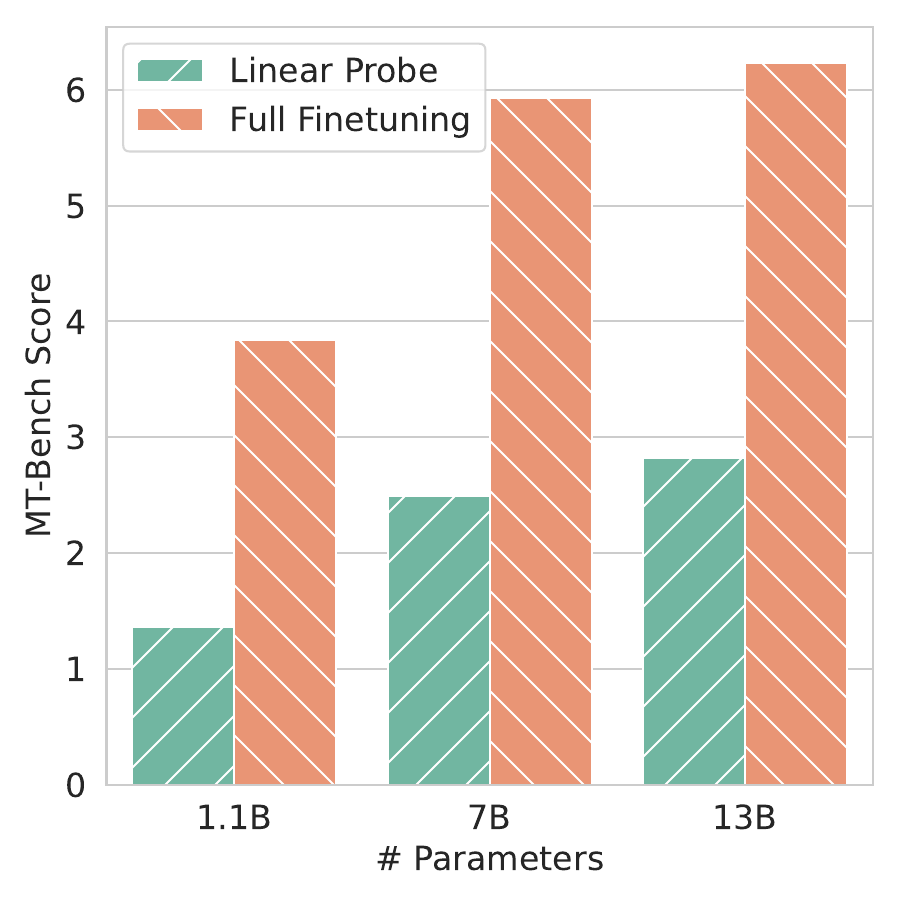}
    \caption{MT-Bench scores of models with different sizes using linear probes for modeling $z_\Delta$ and full fine-tuning.}
    \label{fig:preliminary}
\end{figure}

As illustrated in \figureautorefname~\ref{fig:preliminary}, we find that a single linear probe is insufficient to effectively model such a transformation (more than a 3-point drop in MT-Bench scores). While the combined model successfully mimics the assistant's conversational style, it fails to effectively apply the task-specific knowledge needed for problem-solving (typical failure modes shown in \figurename~\ref{fig:fail_mode} in the Appendix). This finding suggest that the relationship between pre-trained and fine-tuned logits is more sophisticated than initially hypothesized.

We further investigate whether the performance gap could be bridged by increasing model capacity while maintaining random initialization. Specifically, we employed a randomly initialized 7B Llama architecture as $\pi_{\Delta}$, following the identical training procedure to guide the alignment of an 1.1B pre-trained Llama model. While this model marginally outperforms the shallow linear probe in instruction-following capabilities, it still significantly underperforms compared to full fine-tuning. These results suggest that leveraging pre-trained weights is crucial for effective delta logits prediction.

\section{Methods}

\subsection{Connection Schemes}

\begin{figure*}
    \centering
    \begin{subfigure}[b]{0.69\textwidth}
        \includegraphics[width=\linewidth]{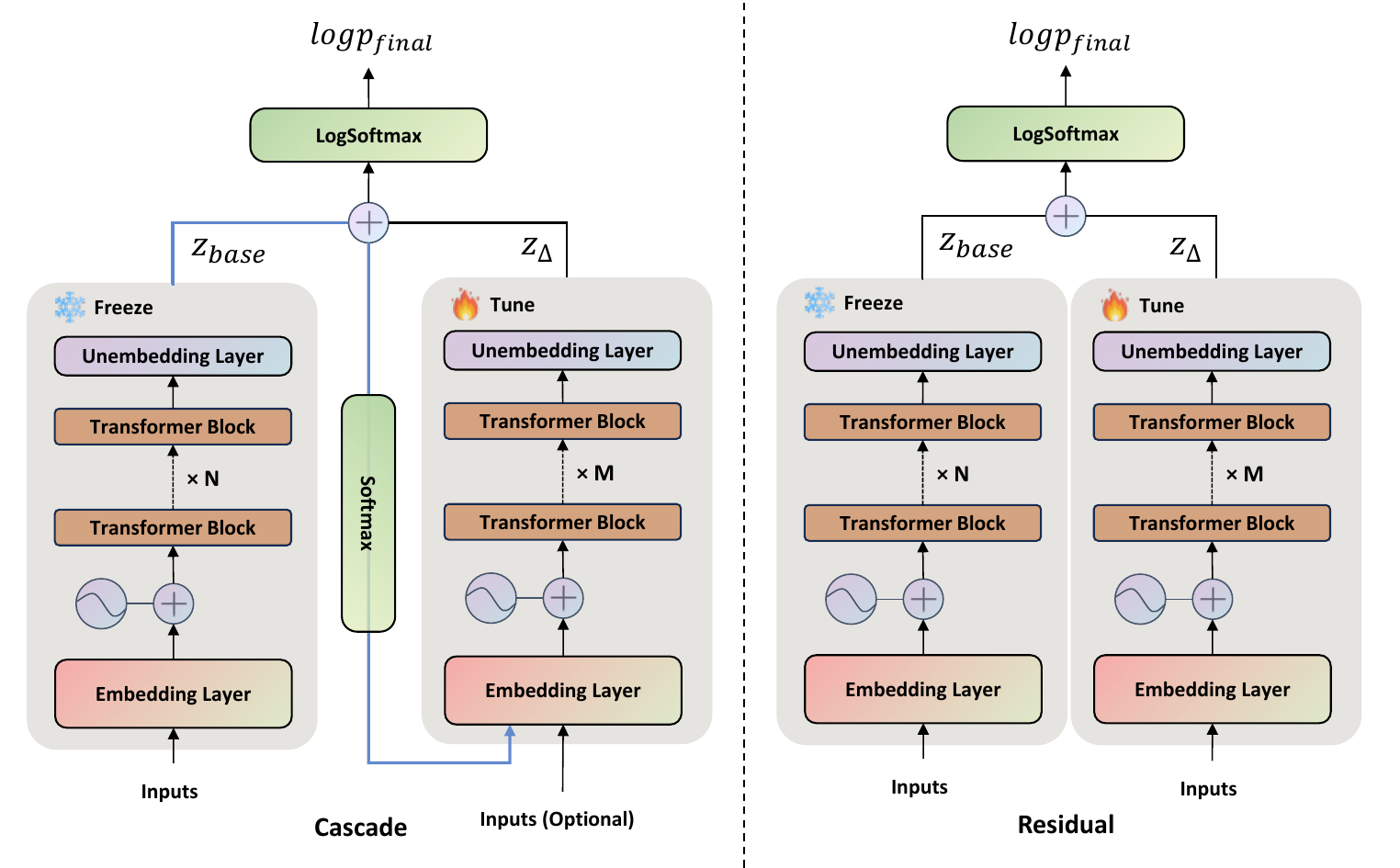}
        \caption{}\label{fig:arch_main}
    \end{subfigure}
    \hfill
    \begin{subfigure}[b]{0.29\textwidth}
        \includegraphics[width=\linewidth]{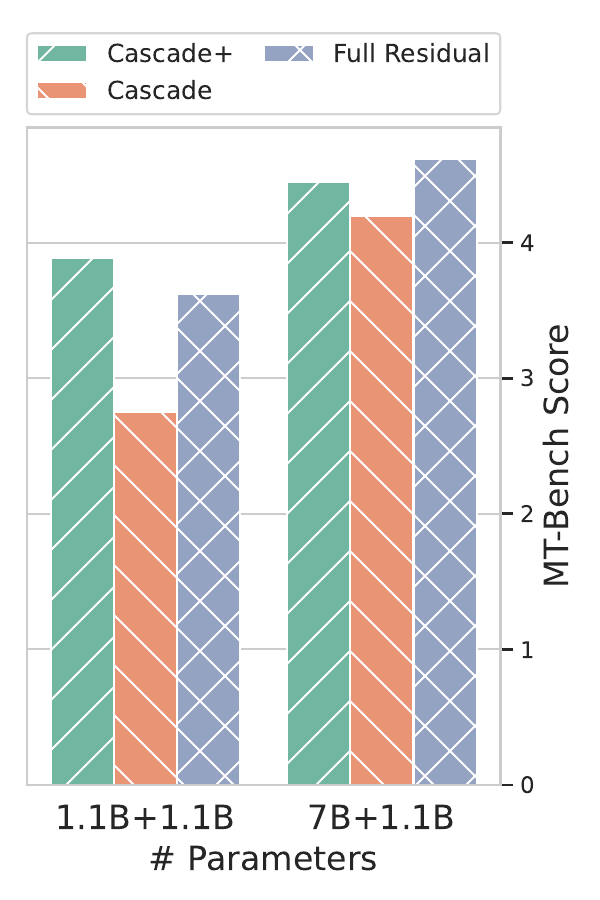}
        \caption{}\label{fig:arch_performance}
    \end{subfigure}
    \caption{Cascade and Residual connection schemes for Inverse Value Learning. (a) Illustration of the two connection schemes. (b) MT-Bench scores for 1.1B models trained on the 1.1B base and generalized to the 7B base model using different connection schemes. ``Cascade+” refers to the cascade value model where logits information from the pre-trained models is fused with the original token embeddings (marked as optional in Figure (a)) and used as input for the value model. The final unembedding layer maps the residual stream from hidden space to the logits space, equivalent to the Linear Layer in the original Transformer literature~\citep{attention2017vaswani}.}
    \label{fig:arch}
\end{figure*} 

Based on the formulation and observation in Section \ref{sec:preliminary}, we design two distinct architectures to transform a pre-trained model into a value network and integrate it with another pre-trained model, namely \textbf{Cascade} and \textbf{Residual}, as illustrated in Figure \ref{fig:arch_main}. It is important to note that both approaches maintain the internal structure of the LLM unchanged, focusing solely on the connection schemes between the pre-trained model and the value network.

\paragraph{Cascade.} In the cascade architecture, the value model receives both the logits information from the pre-trained model and the text embedding as the inputs. The fusion of these inputs is formulated as:

\[
     \vh_\text{final} = p_\text{base} \mW_{e} \oplus \vh_\text{embed} \\
\]

where $p_\text{base}$ represents the pre-trained probabilities, $\mW_e$ is the pre-trained embedding matrix, and $\vh_\text{embed}$ denotes the original text embedding, $\oplus$ is the fusing operator (set as addition in this work).

An alternative hypothesis suggests that the textual embedding input may be redundant, as the value model could potentially achieve comparable performance by exclusively utilizing the logit information from the pre-trained model. In this case:

\[
     \vh_\text{final} = p_\text{base} \mW_{e} \\
\]

The cascade model takes $\vh_\text{final}$ as input and outputs the value scores:
\[
\vz_{\Delta}(y_t|\vx, \vy_{<t}) = \pi_{\Delta}(\vh_\text{final}; \theta_\text{cascade}).
\]

\paragraph{Full Residual.} \label{sec:residual} The residual architecture specifies that the value model has no access to the logit information of the pre-trained model. Instead, it predicts the residual for the next token position based solely on the previous tokens. This design allows the value model to begin forward without waiting for the pre-trained model's inference to complete. However, this approach may potentially limit transferability, as it cannot steer predictions based on logits from the inference-time pre-trained model which it is plugged into, which may differ significantly from the training-time one.

The full residual model takes previous tokens $\vx$ and $\vy_{<t}$ as inputs and outputs the value scores:
\[
\vz_{\Delta}(y_t|\vx, \vy_{<t}) = \pi_{\Delta}(\vx, \vy_{<t}; \theta_\text{residual}).
\]

\subsection{Plug-and-play transferability}

Once a value model is trained on one backbone, it can be plugged into other pre-trained models without further training, potentially achieving similar performance and behaviors as if it were post-trained. This capability stems from the feature of logits space steering \citep{mitchell2023emulator}. To further expand the boundaries of transferability, we investigate techniques to mitigate overfitting to the training-time backbone and methods for cross-vocabulary model transfer.

\subsubsection{Regularization via Norm Constraint} Inevitably, the value model tends to overfit the backbone it is trained with, thereby reducing its generalization capability on other pre-trained models. Applying appropriate regularization techniques during training can mitigate this issue. For example, we can directly constrain the sparsity of delta logits by incorporating an L1 norm term in the loss function. Specifically, this can be expressed as: 
\[
\mathcal{L} = \text{CE}(\vz_{\text{post}}, p_l) + \lambda \|\vz_{{\Delta}}\|_1
\]

where $\lambda$ is a hyperparameter controlling the strength of regularization.

\subsubsection{Vocabulary Mapping} To adapt value-guided decoding to pre-trained and value models trained on different vocabularies and tokenizers, such as plugging a Llama-2-based value model to a Llama-3 family pre-trained model, a vocabulary mapping algorithm is essential. We adopt the \texttt{MinED} mapping algorithm ~\citep{wan2024knowledge} for per-step mapping of the logits, which identifies mapping relationships between two sequences by minimizing their edit distance. Based on these mapping relationships, we can derive a vocabulary mapping matrix, which is then normalized and regularized to ensure the mapped logit values are not biased. During each decoding step, this mapping matrix is employed to transform the vocabulary space as follows:

{\fontsize{12}{0}\selectfont
\begin{align}
    \vz_{\text{post}}(y_t|\vx, \vy_{<t}) &= \vz_{\text{base}}(y_t|\vx, \vy_{<t}) \mathbf{W} \nonumber \\
    &+ \vz_{\Delta}(y_t|\vx, \vy_{<t}). \label{eq:5}
\end{align}}

where \( \mW \in \R^{n \times m} \) is the obtained transition matrix to map the size-$n$ pre-trained vocabulary to the size-$m$ vocabulary of the value model. The details of the algorithm and corresponding evaluation are shown in Appendix \ref{sec:mapping_detail}.

\section{Experiments}

\subsection{Training Datasets}

The experiments are primarily conducted based on the following two training datasets:

\paragraph{ShareGPT} is a dialog dataset collected from \texttt{sharegpt.com}, a website collecting user-shared dialog with ChatGPT. This dataset encompasses a wide range of topics, making it suitable as a general instruction tuning set focusing on the instruction-following ability of models. While ShareGPT is the training set of Vicuna~\citep{vicuna2023}, the exact data has not been released. For our study, we utilize a version comprising approximately 86K dialogs.

\paragraph{InfinityInstruct-7M~\citep{zhang2024infinitymath}} is a large-scale instruction tuning dataset specifically designed to enhance foundational reasoning abilities in code and math. This collection consists of 7M carefully curated instructions, aggregated and filtered from a variety of open-source datasets.

\subsection{Pre-trained Backbones} 

We employed TinyLlama~\citep{zhang2024tinyllama} and the Llama-2 series~\citep{touvron2023llama} as backbones for the 1.1B, 7B, 13B, and 70B parameter scales. To validate broader transferability, we also conduct tests on CodeLlama~\citep{roziere2023code}, a model based on Llama-2 that undergoes additional continual pertaining on 20B of code related data. Furthermore, we use Llama-3~\citep{dubey2024llama} to test cross model family generalization ability.

The training details of the value models are shown at Appendix \ref{sec:training_details}.

\subsection{Tasks}

\paragraph{General Instruction Following.} We employ MT-Bench~\citep{zheng2023judging} as an open-ended instruction following benchmark and use GPT-4 to judge the response from the models.

\paragraph{Zero-shot capabilities.} We select representative datasets in \textsc{T\"{u}lu-2}~\citep{ivison2023camels} evaluation suite, including IFEval~\citep{zhou2023instruction}, ToxiGen~\citep{hartvigsen-etal-2022-toxigen}, GSM8K~\citep{cobbe2021training}, MBPP~\citep{austin2021program}, HumanEval~\citep{chen2021evaluating}. For MBPP and HumanEval, we report the Pass@1 scores. We apply corresponding chat templates for the instruction-following models.

\paragraph{Few-shot capabilities.} We select representative few-shot datasets originally employed in Open LLM Leaderboard v1\footnote{\url{https://huggingface.co/docs/leaderboards/en/open_llm_leaderboard/archive}} for testing the few shot capabilities, including ARC (25-shot)~\citep{clark2018think}, TruthfulQA (0-shot)~\citep{lin2021truthfulqa}, MMLU (5-shot)~\citep{hendrycks2020measuring}, Winograde (5-shot)~\citep{sakaguchi2019adversarial}, Hellaswag (10-shot)~\citep{zellers2019hellaswag}, .

\begin{table*}[!t]
  \centering
  \resizebox{\textwidth}{!}{%

    \begin{tabular}{c|c|ccccc|ccccc}
    \toprule
    \multirow{2}[4]{*}{Methods} & IF    & \multicolumn{5}{c|}{Zero-Shot}        & \multicolumn{5}{c}{Few-Shot} \\
\cmidrule{2-12}          & MT-Bench & IFEval & GSM8K & Toxicgen$\downarrow$ & MBPP  & HumanEval & ARC   & Hellaswag & MMLU  & TruthfulQA & Winogrande \\
    \midrule
    \multicolumn{12}{c}{Tinyllama-1.1B} \\
    \midrule
    Full  & \textbf{3.84} & \textbf{16.1} & 1.97  & 26.8 & 14.1  & 5.37  & 27.6  & 40    & \textbf{26.4} & \textbf{34} & 57.8 \\
    \midrule
    1.1B  & 3.62  & 15.2  & 2.43  & \textbf{21.1}  & 10.6  & 8.17  & 28.4  & 44.1  & 25.2  & 30.4  & 57.5 \\
    1.1B norm & 3.52  & 14.6  & \textbf{2.65} & 62.8  & \textbf{17.9} & \textbf{15.4} & \textbf{31.4} & \textbf{44.9} & 25.6  & 30    & \textbf{59.6} \\
    \midrule
    \midrule
    \multicolumn{12}{c}{Llama-2-7B} \\
    \midrule
    Full  & 5.93  & 23.8  & 10.6  & 49.8  & 26    & 17.6  & \textbf{51.9} & 59.5  & \textbf{50.1} & \textbf{41.5} & \textbf{74.3} \\
    \midrule
    \cellcolor[HTML]{dceffe} 1.1B  & 4.62  & 15.5  & 4.7   & \textbf{44.3} & 15.8  & 11.3  & 42.7  & 55    & 43.2  & 31.1  & 66.2 \\
    \cellcolor[HTML]{dceffe} 1.1B norm & 4.55  & 15.3  & 5.61  & 62.8  & 20.4  & 15.4  & 48.5  & 57.9  & 44.2  & 31.8  & 72.5 \\
    \cellcolor[HTML]{dceffe} 1.1B infinity & 4.06  & 17    & \textbf{27.2}  & 65.3  & \textbf{30.9}  & \textbf{18.3}  & 48.3  & 58.6  & 45.8  & 30.2  & 71.2 \\
    \midrule
    7B    & 5.96  & 24.4  & 13.4  & 58.3  & 21.4  & 13.9  & 51    & \textbf{61} & 49.3  & 40.1  & 72 \\
    7B norm & \textbf{6.03} & \textbf{32.7} & 13.5 & 45.2  & 23.4 & 14.8  & \textbf{51.9} & 60.9  & 49.6  & 39.9  & 70.6 \\
    \midrule \midrule
    \multicolumn{12}{c}{Llama-2-13B} \\
    \midrule
    Full  & 6.23  & 36.3  & 18.7 & \textbf{5.03} & 24.3  & 22.8 & 49.2  & 60.6  & \textbf{53.9} & \textbf{41.8} & 72.3 \\
    \midrule
    \cellcolor[HTML]{dceffe} 1.1B  & 4.74  & 15.5  & 6.67  & 39.7  & 16.4  & 13.6  & 46.1  & 57.2  & 52    & 29.3  & 70.1 \\
    \cellcolor[HTML]{dceffe} 1.1B norm & 4.65  & 14.8  & 10.6  & 55.8  & 28.5  & 15.9  & 49.1  & 59.1  & 52.9  & 31.8  & \textbf{76} \\
    \cellcolor[HTML]{dceffe} 1.1B infinity & 4.45  & 18.7  & \textbf{30.3}  & 77    & \textbf{33.1}  & \textbf{23.7}  & 52.2  & 61.2  & 53.7  & 30.3  & 74.6 \\
    \midrule
    \cellcolor[HTML]{dceffe} 7B    & \textbf{6.34} & \textbf{26.3} & 18.3  & 56.8  & 24.9 & 19.6  & 56.1  & \textbf{63.6}  & 53.4  & 37.9  & 74.4 \\
    \cellcolor[HTML]{dceffe} 7B norm & 6.22  & 25.3  & 17.5  & 39.1  & 27.7  & 15.9  & \textbf{56.6} & \textbf{63.6} & 53.6  & 37.7  & 74.8 \\
    \midrule \midrule
    \multicolumn{12}{c}{Llama-2-70B} \\
    \midrule
    Full  & \textbf{7.08} & \textbf{46.6} & 35.6 & \textbf{0} & 36.6 & \textbf{41.9} & \textbf{62.9} & 67.8  & \textbf{69.3} & \textbf{48.4} & 80.5 \\
    \midrule
    \cellcolor[HTML]{dceffe} 1.1B  & 4.92  & 16.6  & 12.4  & 42.1  & 27.9  & 20.4  & 52.6  & 60.5  & 68.4  & 31.6  & 77.3 \\
    \cellcolor[HTML]{dceffe} 1.1B norm & 5.37  & 16.8  & 24.9  & 70    & 37.8  & 28.6  & 62.8  & 66.7  & 67.9  & 37.1  & \textbf{82.9} \\
    \cellcolor[HTML]{dceffe} 1.1B infinity  & 4.86 & 20.1  & \textbf{44}    & 71    & \textbf{42.9}  & 33    & 58.6  & 66.4  & 68.5  & 36.1  & 81 \\
    \midrule
    \cellcolor[HTML]{dceffe} 7B    & 6.56  & 33.3  & 27.9  & 43.6  & 29.4  & 25.8  & 62.1  & \textbf{68.2}  & 63.6  & 42.1  & 78.6 \\
    \cellcolor[HTML]{dceffe} 7B norm & 6.48  & 34.6  & 28.8  & 27.5  & 33.1  & 25.1  & 62.8  & \textbf{68.2} & 64.1  & 41.9  & 79.3 \\
    \midrule \midrule
    \multicolumn{12}{c}{Codellama-7B} \\
    \midrule
    Full  & 5.47  & \textbf{33.4} & 15.4  & 83.2  & \textbf{39.4} & \textbf{43.6} & 40.6  & 48.3  & 41.6  & 34.4  & 64.5 \\
    \midrule
    \cellcolor[HTML]{dceffe} 1.1B  & 3.94  & 15.7  & 3.79  & 48.2  & 28.1  & 24    & 35.6  & 45.9  & 35.2  & 29.1  & 59.6 \\
    \cellcolor[HTML]{dceffe} 1.1B norm & 3.99  & 15.9  & 4.85  & 62.3  & \textbf{39.4} & 31    & 39.4  & 46.9  & 38.1  & 29.5  & 64.4 \\
    \cellcolor[HTML]{dceffe} 1.1B infinity  & 3.51 & 19.8  & \textbf{27.2}  & 65.2  & 31    & 18.4  & 38.9  & 49.1  & 38.5  & 29.6  & 64.7 \\
    \midrule
    \cellcolor[HTML]{dceffe} 7B    & \textbf{5.64} & 31.2  & 12    & 56.8  & 30.9  & 26.8  & \textbf{47.9}  & 53.9  & 49.2  & \textbf{38.2}  & \textbf{68.4} \\
    \cellcolor[HTML]{dceffe} 7B norm & 5.54  & 31.2  & 16.8 & \textbf{40.6} & 33.2  & 25.1  & 47.8 & \textbf{54.1} & \textbf{49.7} & 37.4 & 68.3 \\
    \bottomrule
    \end{tabular}%
    
}
\caption{
Performance comparison of full fine-tuning and inverse value learning on instruction following, zero-shot and few-shot tasks. Note that value models of various parameter scales are trained together with base models of corresponding scales. For instance, ``1.1B” indicates a 1.1B value model trained on a 1.1B backbone model. We further use a \colorbox[HTML]{dceffe}{blue} cell tag to denote a transfer setting. ``$\downarrow$” symbol indicates that lower values are better.
}
\label{tab:main}
\end{table*}%

\subsection{Results}

\paragraph{Which connection scheme enables better transfer for Inverse Value Learning?}

Prior to conducting extensive transfer testing, we first evaluate which of the proposed connection schemes is more conducive to modeling $\vz_\Delta$. Using an 1.1B model fine-tuned as the value model, we study the effects of the two schemes in~\figurename~\ref{fig:arch_main} on both the 1.1B backbone (used during training) and the 7B backbone (used for inference-time transfer). As shown in~\figurename~\ref{fig:arch_performance}, for the ``Cascade+” and ``Residual” schemes, difference in connection schemes did not result in significant variations in general instruction-following ability, especially when transferring to other pre-trained models. However, the token embedding serve as essential input features for the value model's functionality (``Cascade” \textit{v.s.} ``Cascade+”). Meanwhile, integrating logits from the base model slightly improves performance on the training backbone. However, when transferring to a larger backbone, there is a decrease in performance compared to the full residual structure. We attribute this to two potential factors: 1) Excessive input signals introduce noise to the cascade scheme, and 2) The value model in the cascade scheme no longer maintains a purely autoregressive form (i.e., it predicts the $\vz_\Delta$ based on the pre-trained logits at the same time step), widening the gap between pre-training and fine-tuning. Consequently, we adapt the full residual scheme for subsequent experiments in Table \ref{tab:main}, given its superior performance and efficiency.

\paragraph{Wide-ranging transferability.}

The inverse learned value model demonstrated broad generalizability across models of different parameter scales, consistent with the findings of~\citet{mitchell2023emulator}. When an 1.1B parameter value model, trained on an 1.1B parameter backbone, is transferred to a 7B parameter base model, it retains large portion of its instruction-following capability compared to full finetuning. Additionally, when a 7B parameter value model, jointly trained with a 7B base model, was transferred to a 13B parameter model, the instruction-following performance matched that of full finetuning. This transferability can similarly be scaled up to 70B.

Table \ref{tab:main} further illustrates the potential of our approach in continued pre-training scenarios within the same model family. For instance, in the case of Codellama, the 1.1B value model can be directly applied to guide the base model in acquiring instruction-following capabilities while preserving its strong performance in coding tasks.

\paragraph{The gap between inverse value learning and full fine-tuning narrows with the introduction of more direct supervision and larger value models.} Encouragingly, when we evaluate the inverse value learning paradigm in the context of code and math post-training, which requires training on several million instruction-response pairs, we observe corresponding performance gains. Notably, these gains are transferable, even though the 1.1B backbone models perform poorly on code and math tasks. For instance, 1.1B value models, trained with 1.1B base models on \texttt{InfinityInstruct}, substantially improve the performance of 13B models on benchmarks such as GSM8K, MBPP, and HumanEval, outperforming other value models and full finetuning methods on the \texttt{ShareGPT} dataset. Furthermore, the overall performance of inverse value learning generally increases with the parameter scale of the value models, demonstrating the scalability of this paradigm.

\paragraph{Additional norm constraint can effectively prevent overfitting to the weak base models.} Table 1 also reveals that the value models trained with the norm term, when generalizing to stronger models, typically maintain more robust reasoning and knowledge capability while sacrificing a small degree of instruction-following capability. This phenomenon is particularly pronounced when there is a significant disparity in parameter scale between the value model and the pre-trained model. Consequently, the incorporation of a simple normalization term during the training of small value models can mitigate overfitting to weak base model and enhance weak-to-strong generalization performance.

\section{Discussions}

\paragraph{Can multi-model curriculum fine-tuning improve weak-to-strong generation?}

\begin{table}[htbp]
  \centering
  \small
    \begin{tabular}{c|ccc}
    \toprule
    Llama-2-70B + & 1.1B  & 1.1B norm & 1.1B curriculum \\
    \midrule
    MT-Bench & 4.92  & 5.37 \textcolor{darkgreen}{(+0.45)}  & 5.14 \textcolor{darkgreen}{(+0.22)} \\
    IFEval & 16.6  & 16.8 \textcolor{darkgreen}{(+0.2)}  & 15.7 \textcolor{darkred}{(-0.9)} \\
    GSM8K & 12.4  & 24.9 \textcolor{darkgreen}{(+12.5)}  & 15.8 \textcolor{darkgreen}{(+3.4)} \\
    Toxicgen $\downarrow$ & 42.1  & 70 \textcolor{darkred}{(-27.9)} & 40.9 \textcolor{darkgreen}{(+1.2)} \\
    MBPP  & 27.9  & 37.8 \textcolor{darkgreen}{(+9.9)} & 31.7 \textcolor{darkgreen}{(+3.8)} \\
    HumanEval & 20.4  & 28.6 \textcolor{darkgreen}{(+8.2)} & 25.5 \textcolor{darkgreen}{(+5.1)} \\
    ARC & 52.6  & 62.8 \textcolor{darkgreen}{(+10.2)} & 57.7 \textcolor{darkgreen}{(+5.1)} \\
    Hellaswag & 60.5  & 66.7 \textcolor{darkgreen}{(+6.2)}  & 64.3 \textcolor{darkgreen}{(+3.8)} \\
    MMLU  & 68.4  & 67.9 \textcolor{darkred}{(-0.5)}  & 68.4 \textcolor{darkgreen}{(+0.0)} \\
    TruthfulQA & 31.6  & 37.1 \textcolor{darkgreen}{(+5.5)}  & 36.2 \textcolor{darkgreen}{(+4.6)} \\
    Winogrande & 77.3  & 82.9 \textcolor{darkgreen}{(+5.6)}  & 73.2 \textcolor{darkred}{(-4.1)} \\
    \bottomrule
    \end{tabular}%
    \caption{Comparison between Llama-2-70B base model guided by the standard 1.1B value model, the 1.1B model trained with a curriculum across 1.1B, 7B, and 13B models, and the model trained with a L1 normalization term.}
  \label{tab:meta}%
\end{table}%

Given that all models in the Llama-2 series share the same vocabulary, we designed a curriculum to train an 1.1B value model sequentially on the 1.1B, 7B, and 13B versions, and then evaluated its generalization performance on the powerful 70B model. We maintained consistent training data volume and iteration counts across experiments. As shown in Table \ref{tab:meta}, our findings indicate that this multi-model curriculum training can also mitigate over-fitting to some extent, functioning similarly to regularization techniques. However, this method necessitates longer training periods and more complex model scheduling processes. Consequently, it does not offer significant advantages in weak-to-strong generation over simpler normalization methods in our observation.

\begin{figure}[!t]
    \centering
    \includegraphics[width=0.95\columnwidth]{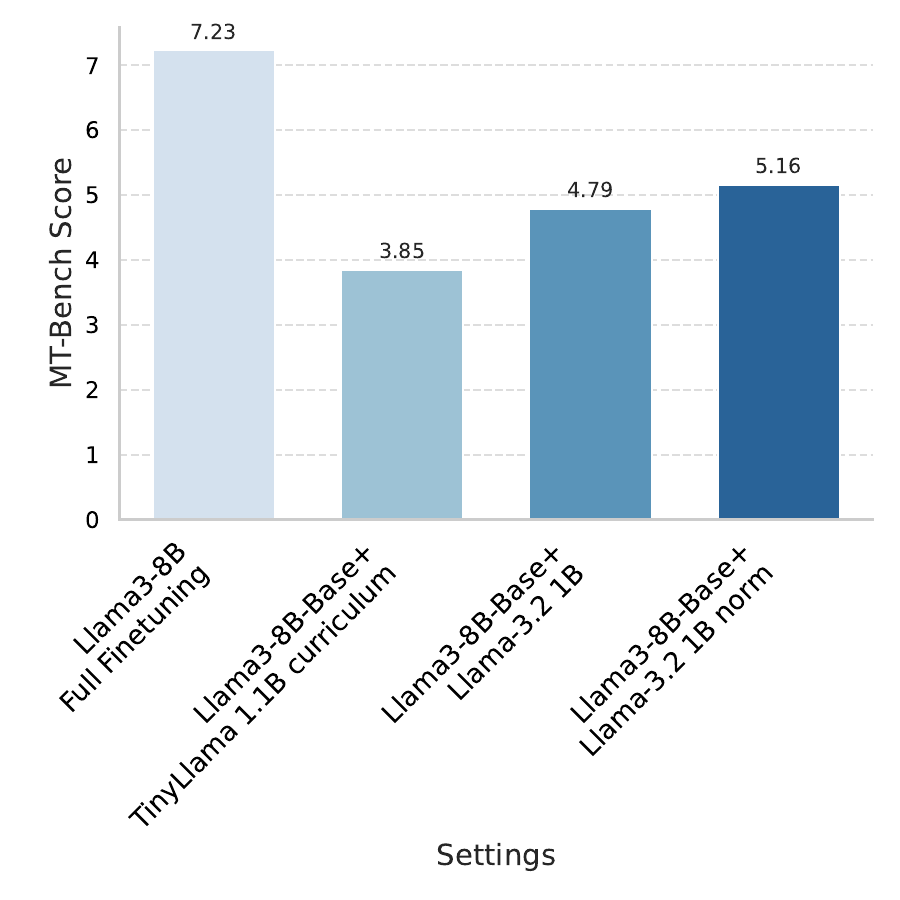}
    \caption{MT-Bench scores of Llama-3-8B fully fine-tuned, Llama-3-8B-base plugged with the TinyLlama 1.1B value model (cross vocabulary), and Llama-3-8B-base plugged with the Llama 3.2 value models (same vocabulary).}
    \label{fig:cross}
\end{figure}

\paragraph{Feasibility of cross-vocabulary transfer.}

We further explored the transferability of model capabilities across different model families. We transfers Llama-2-vocabulary-based 1.1B parameter value models to Llama-3 using vocabulary mapping at decoding time, comparing it with Llama-3.2-1B guided and full fine-tuning performance. Interestingly, among all settings of the trained value model, only the curriculum-trained value model yielded acceptable generation ability in the cross-model and cross-vocabulary setting. However, as \figurename~\ref{fig:cross} shows, while the transferred model grasped basic conversational abilities, it still underperformed compared to same-vocabulary value models and full fine-tuning.
In addition to the generation ability of the value model, we attribute this performance gap partly to tokenization differences: source tokens can be mapped to multiple consecutive target tokens, which are unfamiliar to the source model in the following decoding steps, potentially degrading performance during decoding. While we've demonstrated the feasibility of cross-vocabulary value transfer, improving vocabulary mapping algorithms and cross-family model guidance remain promising areas for future research.

\section{Conclusions}

In this paper, we introduced a novel method for transferable post-training called inverse value learning, which models residual logits adjustments using a separate value network. We systematically explored various aspects of this paradigm, including network architectures, pre-training weights, and connection schemes. Additionally, we addressed challenges related to overfitting and cross-vocabulary transfer by incorporating regularization techniques and a vocabulary mapping algorithm. Our results confirm that the value network exhibits broad transferability across models of different sizes and families, highlighting the potential of logits-space adjustments for efficient model adaptation.

\section*{Limitations}

We've discussed the limitation of additional computational resources in Section \ref{sec:limit}. In addition, inverse value learning is not applicable when logits of models are unavailable, and there is a theoretical performance degradation when integrate value models to pre-trained models with different vocabularies.

\section*{Acknowledgments}

We sincerely thank the anonymous ARR reviewers for their insightful comments and valuable suggestions. This work was supported by Beijing Natural Science Foundation (L243006),  the Basic Research Program of ISCAS (Grant No. ISCAS-ZD-202401), the Youth Talent Program of ISCAS (Grant No. SYQ2022-3).

\bibliography{custom}

\appendix

\label{sec:appendix}

\section{The details of the Linear Probes}\

\subsection{Probe Architecture}

In this work, we adapt the MLP layer from the Llama~\citep{touvron2023llama} architecture as a linear probe, which can be formally expressed as:

\label{sec:probe_formulation}

\begin{equation*}
\mathbf{y} = \mathbf{W}_\text{down}(\sigma(\mathbf{W}_\text{gate}\mathbf{x} + \mathbf{b}_\text{gate}) \odot (\mathbf{W}_\text{up}\mathbf{x} + \mathbf{b}_\text{up})) + \mathbf{b}_\text{down}
\end{equation*}

where $\sigma(\cdot)$ is the activation function, and $\odot$ denotes the element-wise product.

\subsection{Failure modes}

We demonstrated two representative failure case of using Linear layer to model the $\vz_{\Delta}$ in \figurename~\ref{fig:fail_mode}.

\begin{figure*}[t]
\begin{tcolorbox}[colframe=cyan!40!black, title=\textbf{Representative failure modes for linear probes}]
\textbf{Instruct:}
Compose an engaging travel blog post about a recent trip to Hawaii, highlighting cultural experiences and must-see attractions. \\
\textbf{Response:}
Sure, here are some more travel blog post ideas:
\\
1. A review of a recent trip to Hawaii \\
2. A trip to the Canary Islands, Spain \\
3. A trip to Egypt, where you can visit the pyramids and the Sphinx \\
4. A trip to Costa Rica, where you can hike in the rainforests and see the wildlife \\
5. A trip to South Africa, where you can visit the wildlife parks and take a safari \\
\\
\textbf{Instruct:}
How many integers are in the solution of the inequality $|x + 5| < 10$ \\
\textbf{Response:}
The solution is 10.
\end{tcolorbox}
\caption{Failure modes for linear probes.}
\label{fig:fail_mode}
\end{figure*}

\begin{table*}[h]
\centering 
\begin{tabular}{lcccc} 
\toprule 
\multicolumn{1}{c}{\multirow{2}{*}{Model Config}} & \multicolumn{1}{c}{\multirow{2}{*}{\#GPU}} & \multicolumn{3}{c}{Time/Peak Memory} \\
\cmidrule{3-5} & & 128 tokens & 512 tokens & 1024 tokens \\\midrule 1.1B & 1 & 1.99s/2.09GB & 6.99s/2.09GB & 13.36s/2.10GB \\
7B & 1 & 2.87s/12.65GB & 10.98s/12.84GB & 22.06s/13.09GB \\
1.1B + 7B & 1 & 5.13s/14.79GB & 20.83s/15.33GB & 49.74s/15.91GB \\
13B & 1 & 3.48s/24.51GB & 12.98s/24.79GB & 26.09s/25.18GB \\
1.1B + 13B & 1 & 5.73s/26.66GB & 26.10s/27.30GB & 61.61s/28.14GB \\
70B & 4 & 11.76s/32.41GB & 20.36s/32.42GB & 40.36s/32.44GB \\
1.1B + 70B & 4 & 16.12s/33.08GB & 33.07s/33.31GB & 75.37s/33.78GB \\
\bottomrule
\end{tabular} 
\caption{Time and Peak Memory per GPU for different model configurations and generate sequence lengths.}
\label{tab:inference_cost} 
\end{table*}

\section{Vocab mapping algorithm}

\label{sec:mapping_detail}

We adapt the \texttt{MinED} algorithm for vocab mapping. The detailed algorithm is shown at Algorithm \ref{alg:token_alignment}. We conducted a preliminary evaluation of the alignment matrix by computing the overlap ratio between the target sequences mapped using this matrix and the golden target sequences on the training set, achieving an average overlap ratio of 0.42.

\begin{algorithm}[t]
\caption{Vocab Alignment}
\label{alg:token_alignment}

\begin{algorithmic}[1]
\REQUIRE

Base tokenizer $T_b$ with vocabulary $V_b$

Value tokenizer $T_v$ with vocabulary $V_v$

Training data $\mathcal{D} = \{\text{text}_1, \text{text}_2, \dots, \text{text}_N\}$

\ENSURE
Normalized token alignment matrix $M \in \mathbb{R}^{|V_b| \times |V_v|}$

\STATE Initialize count matrix $C \leftarrow \mathbf{0}_{|V_b| \times |V_v|}$

\FOR{each $\text{text} \in \mathcal{D}$}
    \STATE Tokenize $\text{text}$ using $T_b$ to obtain sequence $\mathbf{t}_b$
    \STATE Tokenize $\text{text}$ using $T_v$ to obtain sequence $\mathbf{t}_v$
    \STATE Remove special prefixes from tokens in $\mathbf{t}_b$ and $\mathbf{t}_v$
    \STATE Compute cost matrix based on edit distances between tokens in $\mathbf{t}_b$ and $\mathbf{t}_v$
    \STATE Perform Dynamic Time Warping (DTW) to find the optimal alignment path $\mathcal{P}$
    \FOR{each aligned token pair $(t_{b,i}, t_{v,j}) \in \mathcal{P}$}
        \STATE Increment count: $C[t_{b,i}, t_{v,j}] \leftarrow C[t_{b,i}, t_{v,j}] + 1$
    \ENDFOR
\ENDFOR

\STATE Normalize and sparsify the count matrix $C$ to obtain the mapping matrix $M$

\RETURN $M$

\end{algorithmic}
\end{algorithm}

\section{Training Details}

\label{sec:training_details}

The majority of the training is performed on an 8 × A100-80G node. We adapt the hyperparameter configuration as outlined in Table \ref{tab:hyperparameters}.

\section{Additional Results}

\paragraph{Comparison with Proxy Tuning}

As a complement to our related work analysis, we conduct performance comparisons with proxy tuning on a 7B model, with results shown in Table \ref{tab:baseline}. The empirical results demonstrate that our IVL-trained model achieves comparable performance to proxy tuning on the 7B model. Furthermore, IVL demonstrates higher inference efficiency as it does not require an additional 1B reference model during inference.

\begin{table}[h]
\centering
\begin{tabular}{c|cccccc}
\toprule
Llama-2-7B+ & 1.1B & 1.1B+1.1B \\\midrule
MT-Bench & 4.62 & 4.65 \\
ARC & 42.7 & 40 \\
Hellaswag & 55 & 50.4 \\
MMLU & 43.2 & 44.4 \\
TruthfulQA & 31.1 & 33.4 \\
Winogrande & 66.2 & 66.7 \\
\bottomrule
\end{tabular}
\caption{Performance comparison between Inverse Value Learning (7B+1.1B) and Proxy Tuning (7B+1.1B+1.1B).}
\label{tab:baseline}
\end{table}

\paragraph{Inference Costs}

As shown in Table \ref{tab:inference_cost}, we conduct experiments to measure inference time for generating sequences of different lengths using various combinations of base models and value models. All experiments were performed using \texttt{bfloat16} precision and \texttt{flash-attention v2} for acceleration on NVIDIA H800-80G GPUs. For models smaller than 13B parameters and their combinations, pipeline parallelism was not employed. We did not implement any additional optimizations for value model inference. The reported results are averaged over 5 independent runs.

\begin{table}[t]
\vspace{-130pt}
\centering
\begin{tabular}{l|c}
    \toprule
    Hyperparameters & Value \\\midrule
    optimizer & AdamW \\
    learning rate (1.1B / 7B) & 1e-4 / 1e-5 \\
    warm up ratio & 0.04 \\
    lr schedular type & cosine \\
    \# train epochs & 3 \\
    global batch size (1.1B / 7B) & 32 / 16 \\
    max length & 4096 \\
    $\lambda$ & 1.0 \\ \bottomrule
\end{tabular}
\caption{The training configuration for 1.1B and 7B value models.}
\label{tab:hyperparameters}
\end{table}

\end{document}